\RequirePackage{fix-cm}
\documentclass[smallextended]{svjour3}       
\smartqed  
\usepackage{graphicx} 
\usepackage{amsmath}
\usepackage{amsfonts}
\usepackage[caption=false]{subfig}
\usepackage{algorithm}
\usepackage{algpseudocode}
 \usepackage{natbib}

\usepackage{xcolor}
\usepackage{caption}
\usepackage{bm}

\usepackage[symbol]{footmisc}

\usepackage{xcolor}

\begin{document}







\title{Use of Neural Signals to Evaluate the Quality of Generative Adversarial Network Performance in Facial Image Generation
}

\titlerunning{Neural signals for evaluating GANs}        

\author{Zhengwei Wang$^{*}$ \and Graham Healy\footnote[1]{Equal Contribution}
 \and Alan F. Smeaton \and Tom\'as E. Ward}


\institute{Insight Centre for Data Analytics, Dublin City University, Dublin 9, Ireland \\
            \email{\tt\small zhengwei.wang22@mail.dcu.ie,\tt\small $\left\{\texttt{graham.healy, alan.smeaton, tomas.ward}\right\}$@dcu.ie}           
\textbf{Published at \textit{Cognitive Computation (August 2019)}\\ https://link.springer.com/article/10.1007/s12559-019-09670-y}}

\date{Received: date / Accepted: date}

\maketitle

\vspace{-10pt}
\section*{Abstract}
\vspace{-5pt}
\paragraph{\textbf{Introduction:}}
There is a growing interest in using generative adversarial networks (GANs) to produce image content that is indistinguishable from real images as judged by a typical person. A number of GAN variants for this purpose have been proposed, however, evaluating GANs performance is inherently difficult because current methods for measuring the quality of their output are not  always consistent with what a human perceives. 
\vspace{-10pt}
\paragraph{\textbf{Methods:}} We propose a novel approach that combines a brain-computer interface (BCI) with GANs to generate a measure we call \textbf{Neuroscore}, which closely mirrors the behavioral ground truth measured from participants tasked with discerning real from synthetic images. This technique we call a \textbf{neuro-AI interface}, as it provides an interface between a human's neural systems and an AI process. In this paper, we first compare the three most widely used metrics in the literature for evaluating GANs in terms of visual quality and compare their outputs with human judgments. Secondly we propose and demonstrate a novel approach using neural signals and rapid serial visual presentation (RSVP) that directly measures a human perceptual response to facial production quality, independent of a behavioral response measurement. 
\vspace{-10pt}
\paragraph{\textbf{Results:}} The correlation between our proposed Neuroscore and human perceptual judgments has Pearson correlation statistics: $\mathrm{r}(48) = -0.767, \mathrm{p} = 2.089e-10$. We also present the bootstrap result for the correlation i.e., $\mathrm{p}\leq 0.0001$. Results show that our Neuroscore is more consistent with human judgment compared to the conventional metrics we evaluated. 
\vspace{-10pt}
\paragraph{\textbf{Conclusions:}} We conclude that neural signals have potential applications for high quality, rapid evaluation of GANs in the context of visual image synthesis.

\section{Introduction}
Artificial intelligence (AI) has significant impact on  society yet research into the interaction between humans and AI deserves further exploration and has only recently become a research focus.  
Cognitive computation provides a way of using cognitively inspired techniques to solve a variety of real-world problems and these become especially useful when the interface between an AI system and a human is via a brain-computer interface.
Abbass~\citep{abbass2019social} recently explored the last 50 years of the human-AI relationship with a focus on how the development of trust between the parties has been essential.  He also covered the emergence of direct brain-computer interfaces based on EEG.

\begin{figure}[ht!]
    \centering
    \includegraphics[width=.9\textwidth]{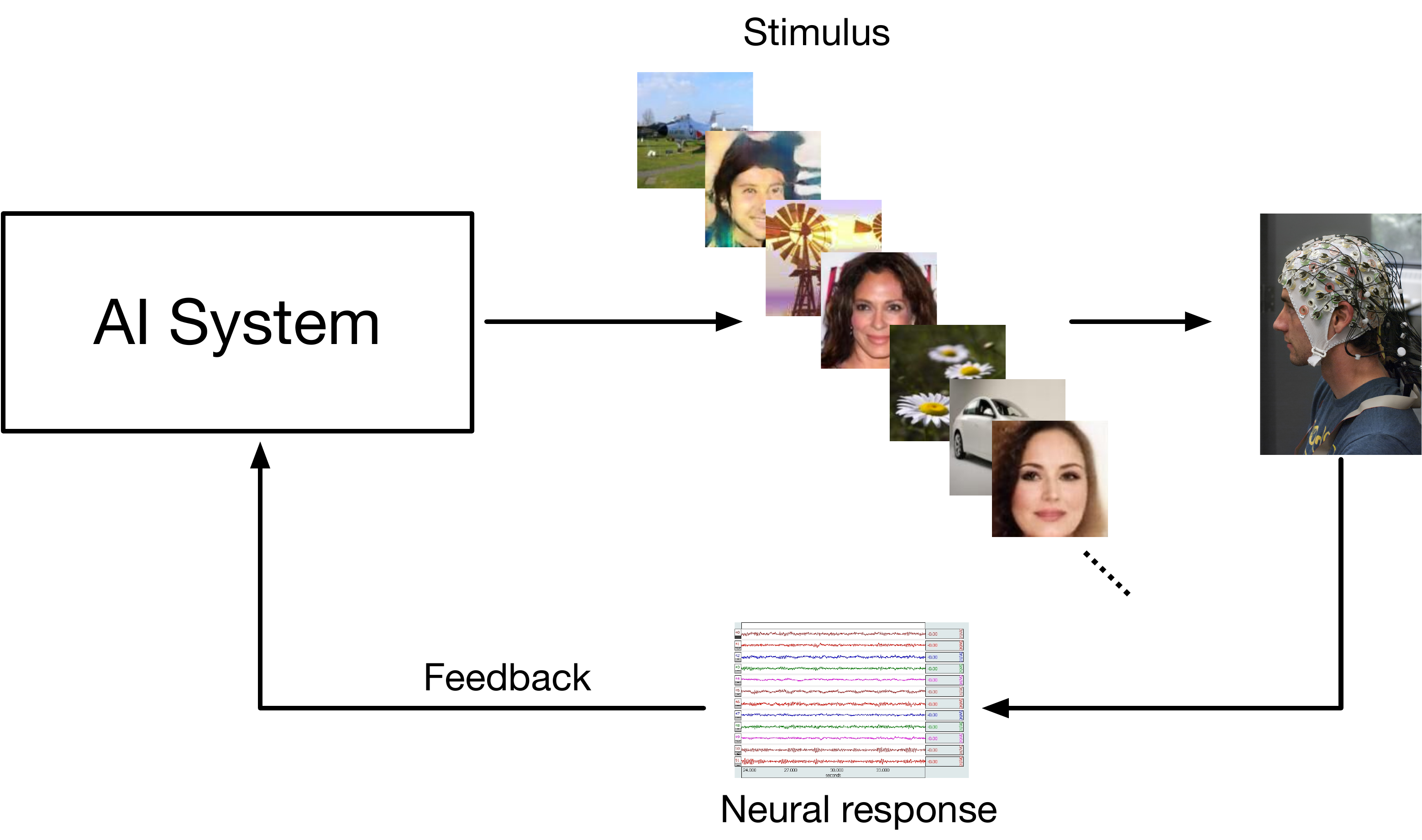}
    \caption{Schematic of neuro-AI interface demonstrated in this study. A type of AI system (e.g., GANs used in this work) produces image stimulus to participants and the corresponding recorded neural response returns to scoring the performance of GANs.}
    \label{fig:neuro-AI-Interface}
\end{figure}

As electroencephalography (EEG) can be the direct reflection of a human's mental processes, the use of EEG is widely studied and deployed in the cognitive computation literature, for example by~\citep{doborjeh2018attentional,li2018hierarchical}. It has been demonstrated recently that EEG can be used effectively for reading emotion~\citep{li2018hierarchical} and that a spiking neural network framework can be used to analyze a human's attention to a task by using EEG~\citep{doborjeh2018attentional}. In this paper, we demonstrate a type of neuro-AI interface derived from cognitive computational perspective (as seen in Fig.~\ref{fig:neuro-AI-Interface}),
which uses neural signals, in this case EEG, to score the performance of generative adversarial networks (GANs).
The relevance between our work and the existing literature such as~\citep{doborjeh2018attentional,li2018hierarchical} is that a processing pipeline has been developed and demonstrated for transforming EEG signals into a value (score or accuracy) and this value matches well a human's cognitive response to a specific class of stimulus, in our case an artificially generated facial image. Moreover, our work contains  experimental details and provides neuroscientific interpretation in the comparison of our EEG-based technique to existing approaches in the literature.

GANs~\citep{goodfellow2014generative} are attracting increasing interests across many different computer vision applications, for example the generation of plausible synthetic images \citep{radford2015unsupervised,arjovsky2017wasserstein,karras2017progressive,berthelot2017began}, image-to-image translation \citep{isola2017image,zhu2017unpaired} and simulated image refinement \citep{shrivastava2017learning}. Despite the extensive work and the many different GAN models reported in the literature, evaluation of the performance of GANs is still challenging. Some comprehensive reviews for GAN evaluation are available including work in \citep{theis2015note,xu2018empirical,borji2018pros} and in summary the evaluation for GANs is divided into  two main types, \textit{qualitative} and \textit{quantitative}. 
The most representative {\it qualitative} metric is to use human annotation to determine the visual quality of the generated images. {\it Quantitative} metrics compare statistical properties between generated and real images. Both approaches have strengths and limitations. 

Qualitative metrics generally focus on how convincing the image is from a human perceptual perspective rather than detecting overfitting, mode dropping and mode collapsing problems \citep{metz2016unrolled}. Human annotation approaches are also time-consuming because they require asking evaluators to generate behavioral responses on an image-by-image basis. 

Quantitative metrics in contrast, are less subjective but the psychoperceptual basis of image quality assessment is not well represented in such metrics hence the robustness of their performance is compromised. As a result, the field of research around evaluation methodologies for GANs is still developing  and presents opportunities for new approaches. One such approach which we propose, is the introduction of a neuro-AI interface, that uses brain signals for image evaluation in the context of a brain-computer interface.

\begin{figure}[ht!]
	\center
	\includegraphics[scale=0.38]{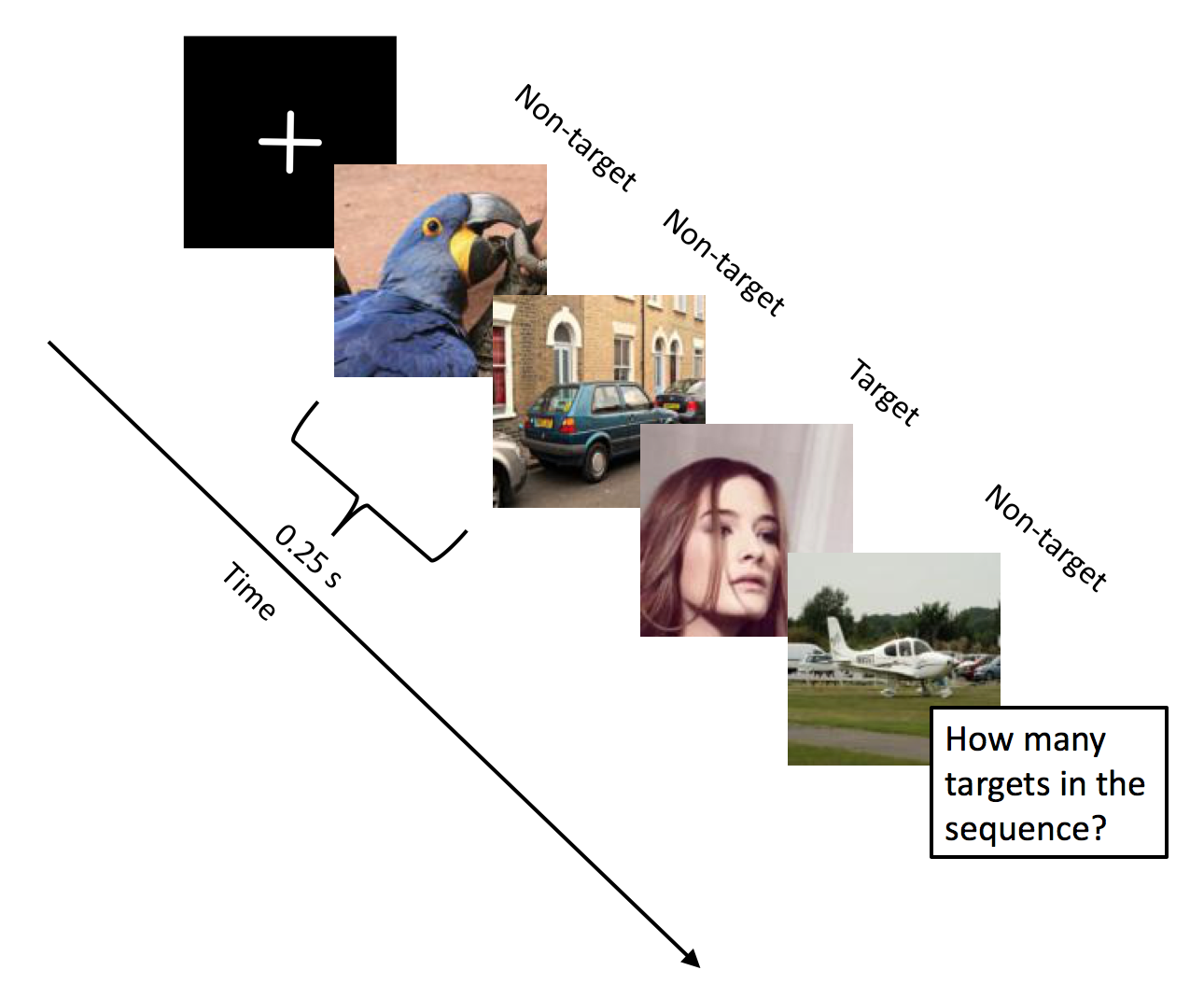}
	\caption{An RSVP image sequence showing juxtaposition of target and non-target images along with a response request.} \label{fig-rsvp_paradigm}
\end{figure}

A brain-computer interface (BCI) is a communication system in which an individual  sends signals to the external world without using the brain's normal output pathways of peripheral nerves and muscles \citep{wolpaw2002brain}. While there are several key BCI applications \citep{lees2018review,healy2017eeg,solon2017deep}, there is a growing interest in using EEG signals in a BCI to help in searching through sets of images. This is based on estimating image content by examining participants' neural signals in response to image presentation. The concept of rapid serial visual presentation (RSVP) can be introduced using a familiar example, that of rapidly riffling through the pages of a book in order to locate a needed image \citep{spence2013rapid}. In RSVP, a rapid succession of target and standard (non-target) images are presented to a participant via a display at a rate of 4 Hz to 10 Hz. The location of target images within the high-speed presentation is not known in advance by participants and hence requires them to actively look out for targets i.e., to attend to target images. This paradigm where participants are instructed to attend to target images amongst a larger proportion of standard images is known as an {\it oddball paradigm} and is commonly used to elicit the P300 event-related potential (ERP), a positive voltage deflection that typically occurs between 300 ms and 600 ms after the appearance of a rare visual target within a sequence of frequent non-relevant stimuli \citep{polich2007updating,hu2010comparison}. Since participants do not know when target images will appear in the presentation sequence, their occurrence causes an attentional-orientation response that is characterized by the presence of a P300 (or P3) ERP. An example of a RSVP paradigm protocol is shown in Fig.~\ref{fig-rsvp_paradigm}
where the participant's task might be to count the number of images with faces, or to recognise the face of a particular individual. 

The P300 ERP can suffer from a low signal-to-noise ratio (SNR) and its appearance spans multiple electrodes on the scalp, which make the precise measurement of P300 activity in the raw, unprocessed EEG epoch difficult. Our previous work~\citep{wang2018review,wang2018spatial} has shown that the P300 can be spatially filtered to improve SNR and reduce dimensionality. The work here will  demonstrate a pipeline that uses LDA beamformer to reconstruct the P300 component for each type of GAN.

Although some work in the GAN evaluation literature has mentioned that quantitative metrics are correlated with human judgment \citep{salimans2016improved,heusel2017gans}, there is no specifically designed work reported  in the literature which compares quantitative metrics with those produced by human judgment. 
It should be noted that  the use of human judgment through annotation to evaluate GANs in terms of visual quality is very effective. 
However, such approaches are very time-consuming and impractical in terms of scale, in real-world applications. Given the advantages of  conventional human annotation approaches, we explore the area of BCI as we know that  neural signals can reflect human perception. In this work, we propose a type of neuro-AI interface for evaluating GAN outputs and we deploy an oddball task for eliciting P300 components via an RSVP protocol, where human subjects are rapidly evaluating images produced by GANs. An evaluation metric called Neuroscore is proposed and the calculation of Neuroscore is demonstrated. Results show this neuro-AI interface is more efficient compared to conventional human annotation approaches and Neuroscore is highly correlated with behavioral human judgment. Given this, our work has two primary contributions:  
\begin{itemize}
\item The design and evaluation of an experiment to compare human assessments with the leading quantitative metrics for GAN performance measurement in terms of image quality. 
\item The demonstration of a fast and efficient neuro-AI interface in which neural signals provide a superior metric for the evaluation of GANs.
\end{itemize}

\section{Preliminaries}
\subsection{Generative Adversarial Networks}
A generative adversarial network (GAN) has two components, the discriminator $D$ and the generator $G$. Given a distribution $\mathrm{\mathbf{x}} \sim p_{\mathrm{\mathbf{x}}}$, $G$ defines a probability distribution $p_{g}$ as the distribution of the samples $G(\mathrm{\mathbf{x}})$. The objective of a GAN is to learn the generator's distribution $p_{g}$ that approximates the real data distribution $p_{r}$. Optimization of a GAN is performed with respect to a joint loss for $D$ and $G$
\begin{equation} \label{GAN_formula}
	\min \limits_{G} \max \limits_{D} \mathbb{E}_{\mathrm{\mathbf{x}} \sim p_{r}} \log[D(\mathrm{\mathbf{x}})] + \mathbb{E}_{\mathrm{\mathbf{x}} \sim p_{\mathrm{\mathbf{x}}}}\log[1 - D(G(\mathrm{\mathbf{x}}))]
\end{equation}
The evaluation of GANs can be considered as an effort to measure the dissimilarity between $p_{r}$ and $p_{g}$. Unfortunately, the accurate estimation of $p_{r}$ is intractable and thus it is not possible to make a good estimation of the correspondence between $p_{r}$ and $p_{g}$. Another challenge for the evaluation of a GAN is how to interpret that the evaluation metric indicates visual quality. Notwithstanding such challenges, metrics are available and we examine three well-known metrics as background and for comparative purposes.

\subsection{GAN evaluation metrics}
This paper uses three of the most widely-used evaluation metrics for GANs in the literature for comparison and we now examine these in turn. 
\subsubsection{Inception Score (IS)} 
The Inception Score is the most widely used GAN performance metric in the literature \citep{salimans2016improved}. It uses a pre-trained Inception network \citep{szegedy2016rethinking} as the image classification model $\mathcal{M}$ to compute
\begin{equation} \label{IS_formula}
	\mathrm{IS} = e^{\mathbb{E}_{\mathrm{\mathbf{x}}\sim p_{g}}[\mathrm{KL}(p_{\mathcal{M}(y \vert \mathrm{\mathbf{x}} )}\vert \vert p_{\mathcal{M}}(y))]}
\end{equation}
where $p_{\mathcal{M}}(y \vert \mathrm{\mathbf{x}})$ is the label distribution of $\mathrm{\mathbf{x}}$ that is predicted by the model $\mathcal{M}$ and $p_{\mathcal{M}}(y)$ is the marginal probability of $p_{\mathcal{M}}(y \vert \mathrm{\mathbf{x}})$ over the probability $p_{g}$. A larger Inception Score will have $p_{\mathcal{M}}(y \vert \mathrm{\mathbf{x}})$ close to a point mass and $p_{\mathcal{M}}(y)$ close to uniform, which indicates that the Inception network is very confident that the image belongs to a particular ImageNet category and all categories are equally represented. A larger Inception Score suggests that the generative model has both high quality and diversity. However, Inception Score may fail in some cases \citep{barratt2018note}. $1/\mathrm{IS}$ ($1/\mathrm{\text{Inception Score}}$) is used as the comparison score in the work in this paper, for consistency with the other two scores examined.

\subsubsection{Kernel Maximum Mean Discrepancy (MMD)}
MMD \citep{gretton2007kernel} is computed as 
\begin{equation} \label{MMD_formula}
	\mathrm{MMD}^{2}(p_{r}, p_{g}) = \mathbb{E}_{ {\mathrm{\mathbf{x}}_{r}, \mathrm{\mathbf{x}}_{r}^\top,  \sim p_{r} }, \atop {\mathrm{\mathbf{x}}_{g}, \mathrm{\mathbf{x}}_{g}^\top \sim p_{g} } } \\
	 {[k(\mathrm{\mathbf{x}}_{r}, \mathrm{\mathbf{x}}_{r}^\top) - 2k(\mathrm{\mathbf{x}}_{r}, \mathrm{\mathbf{x}}_{g}) + k(\mathrm{\mathbf{x}}_{g}, \mathrm{\mathbf{x}}_{g}^\top)]}
\end{equation}
It measures the dissimilarity between $p_{r}$ and $p_{g}$ for some fixed kernel function $k$. A Gaussian kernel, defined as $k(\mathrm{\mathbf{x}}, \mathrm{\mathbf{x}}^\top) = e^{ -\frac{\vert \mathrm{\mathbf{x}} - \mathrm{\mathbf{x}}^\top \vert ^{2}}{2\sigma}  }$ where $\mathrm{\mathbf{x}}$ are input samples and $\sigma$ is the bandwidth parameter, is often used for this purpose \citep{li2015generative}. A lower MMD indicates that $p_{g}$ is closer to $p_{r}$, indicating a GAN has better performance.

\subsubsection{The Frechet Inception Distance (FID)}
FID~\citep{heusel2017gans} uses a feature space extracted from a set of generated image samples by a specific layer of the Inception network. Regarding the feature space as multivariate Gaussian, the mean and covariance are estimated for both the generated data and real data. FID is computed as
\begin{equation} \label{FID_formula}
	\mathrm{FID}(p_{r}, p_{g}) = \vert \vert \mathrm{\pmb{\mu}}_{r} - \mathrm{\pmb{\mu}}_{g} \vert \vert_{2}^{2} + \mathrm{Tr}(\mathrm{\mathbf{\Sigma}}_{r} + \mathrm{\mathbf{\Sigma}}_{g} - 2(\mathrm{\mathbf{\Sigma}}_{r}\mathrm{\mathbf{\Sigma}}_{g})^{\frac{1}{2}})
\end{equation}
Similar to MMD, a smaller FID indicates better GAN performance.

\subsubsection{Comparing metrics}
In the case of the Inception Score, this is calculated through the Inception model~\citep{szegedy2016rethinking}. It has been shown previously that Inception Score is very sensitive to the model parameters~\citep{barratt2018note}. Even scores produced by the same model trained using different libraries (e.g., Tensorflow, Keras and PyTorch) differ a lot from each other. Inception Score also requires a large sample size for the accurate estimation of $p_{\mathcal{M}}(\mathrm{y})$. FID and MMD both measure the similarity between training images and generated images based on the feature space~\citep{xu2018empirical}, since the pixel representations of images do not naturally support the computation of  meaningful Euclidean distances~\citep{forsyth2003modern}. The main concern about the FID and MMD methods is whether the distributional characteristics of the feature space exactly reflect the distribution for the images~\citep{forsyth2003modern}.

In general, these conventional metrics are easily affected by small artefacts in either pixel space or feature space. For instance, some sharp artefacts in BEGAN may cause large difference between real and generated images regarding the distribution. However, such sharp artefacts would not affect image content and quality as human perception is more robust to conventional metrics regarding these issues.

\subsection{The Event-related potential and P300 (or P3) component}
\begin{figure}[ht!] 
\center
	\includegraphics[scale=0.38]{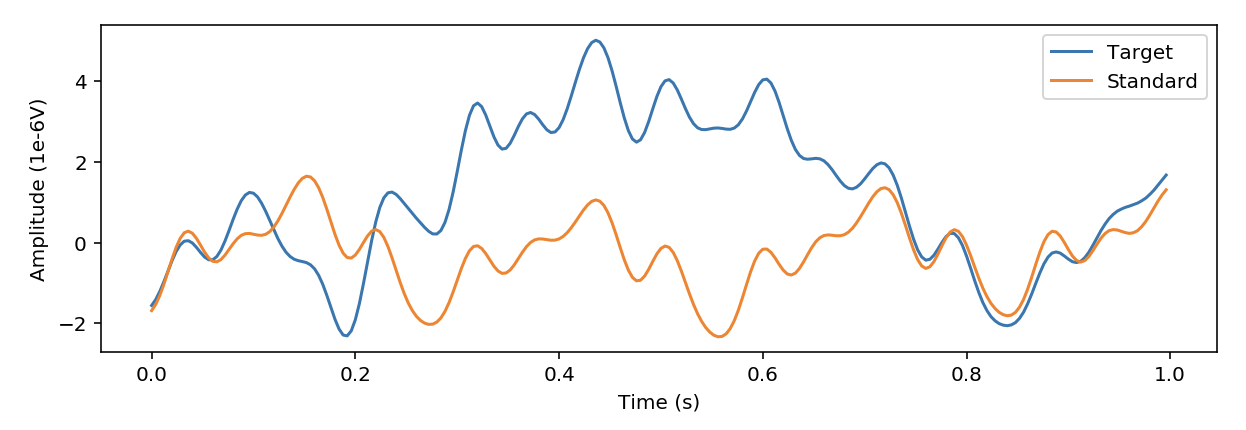} 
	\label{fig-P300_example-a}
\caption{Averaged ERP response for \textit{Participant 9} showing P300-related activity.} \label{fig-P300_example}
\end{figure}

In neuroscience, event-related potentials (ERPs) refer to low amplitude voltage signals measured on the scalp which arise from current source dynamics in the brain whose changes reflect specific events or stimuli~\citep{blackwood1990cognitive}. ERPs are characterized by EEG changes that are time-locked to sensory, motor or cognitive events, and provide a safe and non-invasive approach to study psychophysiological correlates of mental processes \citep{sur2009event}. ERPs can be elicited by a wide variety of sensory, cognitive or motor events. The P300 ERP component was discovered by Sutton \citep{sutton1965evoked} and since then has been one of the most investigated ERP components. The P300 can be elicited when a participant is instructed to respond mentally or physically to a target stimulus and not respond otherwise in the experiment. In this way, it reflects a participant's attention, that is it can be modulated by the specific instruction given to a participant. Figure~\ref{fig-P300_example}
shows an averaged P300 response elicited by a target stimulus that is typically evident between 300 ms and 600 ms post presentation of a stimulus, depending on the type of task. A list of related physiologically-relevant terminology and associated explanations used in this work is presented below: 
\begin{itemize}
\item \textit{Trial:} Each individual image presentation is called a trial.

\item \textit{Epoch:}
An epoch is a specific time window which is extracted from the continuous EEG signal. Each epoch is time-locked with respect to an event (image stimulus presentation in our case).

\item \textit{Single trial P300 amplitude:} This is the amplitude of the P300 component corresponding to each individual image. The P300 amplitude is calculated by selecting the maximum voltage value between 400 ms and 600 ms for each EEG epoch.

\item \textit{Averaged P300 amplitude:} This is the difference between the averaged target (for example a face) trial amplitudes and the averaged standard trial amplitudes (for example a non-face).

\item \textit{Reconstructed single trial P300 amplitude:} This is the P300 amplitude corresponding to each single target image. It is the LDA-beamformed single trial P300 amplitude (the detail of the LDA beamformer method is introduced in later in Section~\ref{sec:P300-reconstruction}).

\item \textit {Reconstructed averaged P300 amplitude:} It is the difference between the averaged LDA beamformed P300 corresponding to target trials and the averaged LDA beamformed signal corresponding to standard trials (non-face). 

\end{itemize}

\section{Methodology}
\subsection{Data acquisition and experiment}
\begin{figure}[ht!]
\center
\subfloat{
	\includegraphics[width=.2\linewidth]{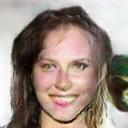} 
    \hspace{5pt}
    \includegraphics[width=.2\linewidth]{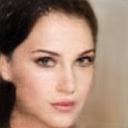} 
    \hspace{5pt}
    \includegraphics[width=.2\linewidth]{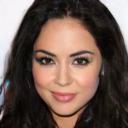} 
    \hspace{5pt}
    \includegraphics[width=.2\linewidth]{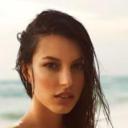} 
}

\caption{Face image examples used in the experiment. From left to right: DCGAN, BEGAN, PROGAN, and real face (RFACE). Images used in the experiment are $128 \times 128$.} \label{fig-face-example}

\end{figure}

We used three GAN models to generate synthetic images of faces: DCGAN~\citep{radford2015unsupervised}, BEGAN~\citep{berthelot2017began} and progressive growing of GANs (PROGAN)~\citep{karras2017progressive} as shown in Fig.~\ref{fig-face-example}.
Image streams in the experiment contain generated images from DCGAN, BEGAN and PROGAN, as well as real face (RFACE) images and non-face category images. RFACE images were sampled from CelebA dataset~\citep{liu2015faceattributes}. Non-face category (standard images) were sampled from ImageNet dataset~\citep{deng2009imagenet}, similar to those used in other RSVP experiments such as~\citep{healy2017eeg,nails}.

EEG data for 12 participants was gathered. Data collection was carried out with approval from Dublin City University Research Ethics Committee (REC/2018/115). Each participant completed two types of tasks which we call the behavioral experiment (BE) task and the rapid serial visual presentation (RSVP) task. The sequence of blocks presented in the experiment was: BE $\rightarrow$ RSVP $\rightarrow$ BE $\rightarrow$ RSVP $\rightarrow$ BE.

The objective of the BE task was to record participants' responses to each type of image category while  the RSVP task was to record EEG when participants were seeing the rapid presentation of images. The ultimate goal of this study was to compare whether the EEG responses in the RSVP task were consistent with  participants' responses in the BE task.

The BE task consisted of three blocks, where each block contained 90 images (18 images for each face category resulting in 72 face images in total and 18 non-face images). Thus there were 216 face images and 54 non-face images in the BE task in total. Participants were presented with one image at a time and asked to press a button corresponding to a ``Yes" if they perceived a real face (i.e., belonging to the real face (RFACE) set) or a ``No" for anything they perceived as not being a real face (including fake face produced by GANs and non-face). Following each response, feedback was given on whether or not the presented image was indeed a real face to make participants pay more attention to the task. The accuracy (number of correct trials divided by number of presented images for that GAN type) of each participant's responses was recorded and their performance is referred to subsequently as a ``human judgment" metric.

The RSVP task contained 26 blocks. Each RSVP block contained 240 images (6 images for each face category thus 24 face targets in total and 216 non-face images), thus there were 6,240 images (624 face targets/5,616 non-face images) available for each participant. In the RSVP task, image streams were presented to participants at a 4 Hz presentation rate. Participants were asked to search for real face (RFACE) images in this task so as to elicit a P300. 
We compare the P300 amplitude in the RSVP task to the human judgment measure in the BE task to determine if they are consistent with each other. 

EEG was recorded for both of the BE and RSVP tasks along with timestamping information for image presentation and behavioural responses (via a photodiode and hardware trigger) to allow for precise epoching of the EEG signals for each trial \citep{wang2016investigation}. EEG data was acquired using a 32-channel BrainVision actiCHamp at 1,000 Hz sampling frequency, using electrode locations as defined by the International 10-20 system. 

To enhance the low signal-to-noise ratio of the acquired EEG, pre-processing is required. Pre-processing typically involves re-referencing, filtering the signal (by applying a bandpass filter to remove environmental noise or to remove activity in non-relevant frequencies), epoching (extracting a time epoch typically surrounding the stimulus onset) and trial/channel rejection (to remove those containing artifacts). In this work, a common average reference (CAR) was utilized and a bandpass filter (i.e., 0.5-20 Hz) was applied prior to epoching. EEG data was then downsampled to 250 Hz. Only behavioral responses occurring between 0 and 1 second after the presentation of a stimulus were used. Trial rejection was carried out to remove those trials containing noise such as eye-related artifacts (via a peak-to-peak amplitude threshold across all electrodes). Details of the retained trials for each participant are shown in Table~\ref{tab:image-number}. 
\begin{table}[ht!]
    \centering
    \begin{tabular}{|c|c|c|c|c|c|}
        \hline
        {\textbf{\textit{ID}}} & {\textbf{\textit{DCGAN}}} & {\textbf{\textit{BEGAN}}} & {\textbf{\textit{PROGAN}}} & {\textbf{\textit{RFACE}}} & {\textbf{\textit{Standard}}} \\ \hline \hline
        {1} & {116} & {108} & {107} & {113} & {4,220}   \\ \hline
        {2} & {100} & {106} & {110} & {98} & {3,215}   \\ \hline
        {3} & {156} & {153} & {154} & {154} & {5,553}   \\ \hline
        {4} & {144} & {153} & {143} & {144} & {5,168}   \\ \hline
        {5} & {110} & {101} & {92} & {80} & {4,150}   \\ \hline
        {6} & {135} & {131} & {122} & {106} & {4,521}   \\ \hline
        {7} & {138} & {139} & {143} & {141} & {4,955}   \\ \hline
        {8} & {151} & {151} & {150} & {151} & {5,290}   \\ \hline
        {9} & {146} & {149} & {140} & {149} & {4,832}   \\ \hline
        {10} & {104} & {87} & {93} & {82} & {3,286}   \\ \hline
        {11} & {149} & {138} & {144} & {142} & {5,270}   \\ \hline
        {12} & {97} & {92} & {99} & {101} & {3,859}   \\ \hline 
    \end{tabular}
    \caption{Number of trials for each stimulus type remaining after artifact rejection.}
    \label{tab:image-number}
\end{table}
A LDA beamformer \citep{treder2016lda} was applied to the retained EEG epochs for each participant to enhance the signal-to-noise ratio (SNR). Details of the application of the LDA beamformer method is described in Section~\ref{sec:P300-reconstruction}.

\subsection{P300 reconstruction} \label{sec:P300-reconstruction}
EEG in our study was recorded using a number of spatially distributed electrodes across the scalp (32 channels of EEG in this study). The P300 is typically predominant on the posterior electrodes of the scalp, which also means the P300 is detected in multiple channels simultaneously. We use the LDA beamformer \citep{treder2016lda} to reconstruct the P300 in this work for the following reasons. Firstly, it is difficult to compare P300 between participants across a number of channels as the location of the P300 varies across participants. Secondly, the P300 suffers from interference from strong background brain activity so it has a very low signal-to-noise ratio (SNR) \citep{luck2014introduction}. The LDA beamformer method allows us to reconstruct the P300 from a multi-dimensional set of EEG signals i.e., transform 32 channels of EEG to a one-channel time series facilitating within-subject comparisons (with the additional benefit of improving the SNR for the reconstructed P300 as well). Given an EEG epoch $\mathbf{X}_{i} \in \mathbb{R}^{C \times T}$ ($C$ is the number of channels and $T$ is time series points in that EEG epoch), let $\mathbf{p}_{1} \in \mathbb{R}^{C\times1}$ and $\mathbf{p}_{2} \in \mathbb{R}^{C\times1}$ be the spatial patterns of a particular component in two different experimental conditions, e.g., face stimuli versus non-face stimuli in this paradigm. We denote the difference pattern as $\mathbf{p} := \mathbf{p}_{1}-\mathbf{p}_{2}$ and the covariance matrix as $\mathrm{\mathbf{\Sigma}} \in \mathbb{R}^{C \times C}$~\citep{treder2016lda}. The optimization problem for the LDA beamformer is to find a projection vector (we call it a spatial filter in the area of EEG/BCI) $\mathbf{w} \in \mathbb{R}^{C \times 1}$ that satisfies

\begin{equation} \label{beam_cost_function}
	\min\limits_{\mathbf{w}} \mathbf{w}^\top \mathrm{\mathbf{\Sigma}} \mathbf{w} \hspace{5pt} s.t. \mathbf{w}^\top\mathbf{p}=1 
\end{equation}
The optimal projection vector $\mathbf{w}$ (in equation \ref{beam_cost_function}) can be calculated as

\begin{equation}\label{eq:w}
	\mathbf{w}=\mathrm{\mathbf{\Sigma}}^{-1} \mathbf{p} ( \mathbf{p}^\top \mathbf{\Sigma}^{-1} \mathbf{p}) ^{-1}
\end{equation}
After determining the optimal $\mathbf{w}$, a high dimensional EEG epoch then can be projected to the one dimensional subspace (reconstructed signal) as

\begin{equation}
	\mathbf{S}_{i} = \mathbf{w}^\top \mathbf{X}_{i}
\end{equation}
where $\mathbf{S}_{i} \in \mathbb{R}^{1 \times T}$ is one trial reconstructed source signal. The LDA beamformer method can be applied to different time regions to reconstruct different individualized spatial profiles for ERP components present in that time frame~\citep{wang2018spatial}. In this study, we apply the LDA beamformer between 400 ms and 600 ms in order to best extract the P300. 

\subsection{Neuroscore}
The \textit{reconstructed averaged P300 amplitude} is used as the basis for our novel metric for evaluating  GAN outputs. To address latency of the P300 which varies across participants, this work \citep{wang2018spatial} has successfully demonstrated the use of LDA beamformer to search for the optimal P300 time index in an RSVP experiment. We select the maximum value in the 200 ms time window which is centered at the optimal time index to represent the \textit{reconstructed single trial P300 amplitude} and then average these across the trials to get the \textit{reconstructed averaged P300 amplitude}. This reconstructed averaged P300 amplitude is the Neuroscore. The process of calculating Neuroscore can be seen in the algorithmic block below. 

\begin{algorithm}[!htbp]
    \caption{Steps for calculating Neuroscore}
    \label{al:calculate-NS}
    \begin{algorithmic}[1]
    \item[\textbf{Input:}]
    \begin{itemize}
        \Statex \item $\mathbf{X} \in \mathbb{R}^{N \times C \times T}$ is the EEG corresponding to target stimulus, $N$ is the number of target trials, $C$ is number of channels, $T$ is number of time points.
    	\item $\mathbf{K} \in \mathbb{R}^{M \times C \times T}$ is the EEG corresponding to standard stimulus, $M$ is number of standard trials, $C$ is number of channels, $T$ is number of time points.
    \end{itemize}
    \item[\textbf{Output:}] Neuroscore
    \State $\mathbf{\Sigma} = \frac{1}{N} \sum_{i=1}^{N}\mathbf{X}_{i}{\mathbf{X}_{i}}^\top + \frac{1}{M} \sum_{i=1}^{M}\mathbf{K}_{i}{\mathbf{K}_{i}}^\top$
    \For{$t_{i}$ in [400 ms, 600 ms]}
        \State $\mathbf{p}=\frac{1}{N}\sum_{i=1}^{N}\mathbf{X}_{i, t_{i}} - \frac{1}{M} \sum_{i=1}^{M}\mathbf{K}_{i, t_{i}}$
        \State {$\mathbf{w}=\mathbf{\Sigma}^{-1} \mathbf{p} ( \mathbf{p}^{\top} \mathbf{\Sigma}^{-1} \mathbf{p}) ^{-1}$}
        \State {$\mathrm{J}_{t_{i}} \gets \bm{\mathrm{w}}^\top \bm{\mathrm{\Sigma}} \bm{\mathrm{w}}$}
        \State{$\mathrm{W}_{t_{i}} \gets \bm{\mathrm{w}}$}
    \EndFor
    \State $t_{optimal}$=$\mathrm{argmin}_{t_{i}} \mathrm{J}$
    \State $\bm{\mathrm{w}}_{optimal}$=$\mathrm{W}_{t_{optimal}}$
    \State $\mathrm{t}_{P300}$=[$\mathrm{t}_{optimal}$ - $100$ ms, $\mathrm{t}_{optimal}$ + $100$ ms] \Comment \textit{This is time window being detected for P300.} 
    \For{$i = 1:N$}
        \State{$\mathbf{s} = \mathbf{w}^\top\mathbf{X}_{i}$}
        \State{$a = \max(\mathbf{s}_{t_{P300}}$)}
        \State{$\mathrm{A}_{i} \gets$ a}
    \EndFor
    \State $\mathrm{Neuroscore}$ = $\dfrac{1}{N}\sum_{i=1}^{N}\mathrm{A}_{i}$
    \end{algorithmic}
\end{algorithm}

\noindent 
It should be noted that Neuroscore benefits from a high SNR compared to the traditional single trial P300 for the following reasons: 
\begin{enumerate}
    \item The LDA beamformer has been applied to raw EEG epoch data in order to maximize the SNR;
    \item Neuroscore is calculated by averaging trials which is able to mitigate the background EEG noise.
\end{enumerate} 
Hence, our proposed Neuroscore is a relatively robust metric as defined for this work. It should be noted that higher Neuroscore values indicate better GAN performance which is inverse to the traditional scores used in this work.


\section{Experimental Results}
\subsection{Behavior task performance}
We included 12 participants in the BE tasks and recorded the accuracy (calculated as the number of correctly labelled images divided by the total number of images) of their judgments for each face category. In Table~\ref{BE-ACC}
\begin{table}[!htp] 
	\centering
	\begin{tabular}{|c||c|c|c||c|}
		\hline
		{\textbf{\textit{ID}}}& {\textbf{\textit{DCGAN}}} & {\textbf{\textit{BEGAN}}} & {\textbf{\textit{PROGAN}}} & {\textbf{\textit{RFACE}}}  \\
		\hline \hline
		{1} &{1.000} & {0.759} & {0.704} & {0.759}  \\
		\hline
		{2} &{0.981} & {0.741} & {0.537} & {0.537} \\
		\hline
		{3} &{1.000} & {0.796} & {0.778} & {0.537} \\
		\hline
		{4} &{0.981} & {0.889}& {0.704}  & {0.667} \\
		\hline
		{5} &{1.000} & {0.667} & {0.648}  & {0.759}\\
		\hline
		{6} &{1.000} & {0.926}& {0.704}  & {0.759} \\
		\hline
		{7} &{1.000} & {0.815}& {0.611}  & {0.759} \\
		\hline
		{8} &{0.981} & {0.815} & {0.870}  & {0.759}\\
		\hline
		{9} &{1.000} & {0.796}  & {0.685} & {0.704}\\
		\hline
		{10} &{1.000} & {0.815}  & {0.759} & {0.722}\\
		\hline		
		{11} &{1.000} & {0.907}  & {0.759} & {0.685}\\
		\hline
		{12} &{1.000} & {0.963}  & {0.704} & {0.796}\\
		\hline		\hline 
		{Mean} &{\textbf{0.995}} & {0.824} & {\textbf{0.705}} & {0.695} \\
		\hline
	\end{tabular}
	\caption{Accuracy for face images generated from three GANs and real face images in the BE task. Lower accuracy for GAN-generated images indicates better image quality  i.e.,  participants were often convinced that synthesised faces were in fact real.}
	\label{BE-ACC}
\end{table} 
it can be seen that participants achieve the lowest accuracy (0.705) for PROGAN and the highest accuracy (0.994) for DCGAN i.e., participants rank PROGAN, BEGAN and DCGAN from high performance to low performance respectively. While learning effects may be present,  our result is robust regardless of  learning effects as we examined using different groups of RSVP blocks combined with different parts of the BE task, and the results remained consistent. It is interesting that human judgment accuracy for RFACE is 0.695 which is comparatively low. This may be caused by participants being convinced by GAN generated images and subsequently feeling less confident on  RFACE images, which indicates that GANs are able to convince participants in this case.

\subsection{Rapid Serial Visual Presentation task performance} \label{sec:rsvp-performance}
In order to employ neural signals to evaluate the performance of GANs, we use the RSVP paradigm to elicit the P300 ERP.  
\begin{figure}[!htbp] 
\begin{center}
	\includegraphics[scale=0.7]{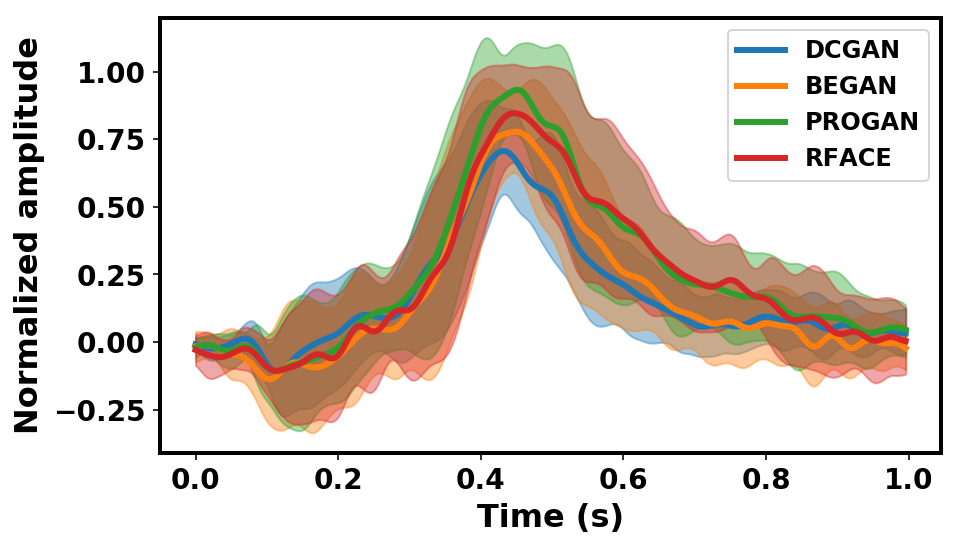}
	\caption{Reconstructed averaged (via LDA beamformer) P300 signal across 12 participants in this study.} \label{fig-RSVP-P300}
\end{center}
\end{figure}
Figure~\ref{fig-RSVP-P300} shows the \textit{reconstructed averaged P300 signal} across all participants (using LDA beamformer) in the RSVP experiment. It should be noted here that the \textit{reconstructed averaged P300 signal} is calculated as the difference between averaged target trials and averaged standard trials after applying the LDA beamformer method i.e., $\frac{1}{n} \sum \limits_{i=1}^{n} \mathbf{w}^\top\mathbf{X}^{target}_{i} -  \frac{1}{m} \sum \limits_{i=1}^{m} \mathbf{w}^\top\mathbf{X}^{standard}_{i} $, where $\mathbf{w}$ is the spatial filter calculated by LDA beamformer, $\mathbf{X}$ are the EEG epochs, $n$ and $m$ are the numbers of targets and standards respectively. The solid lines in Figure~\ref{fig-RSVP-P300} are the means of the reconstructed averaged P300 signals for each image category (across 12 participants) while the shaded areas represent the standard deviations (across participants). It can be seen that the reconstructed averaged P300 (across participants) clearly distinguishes between different image categories. 

\begin{figure}[!htbp] 
\begin{center}
	\includegraphics[scale=0.15]{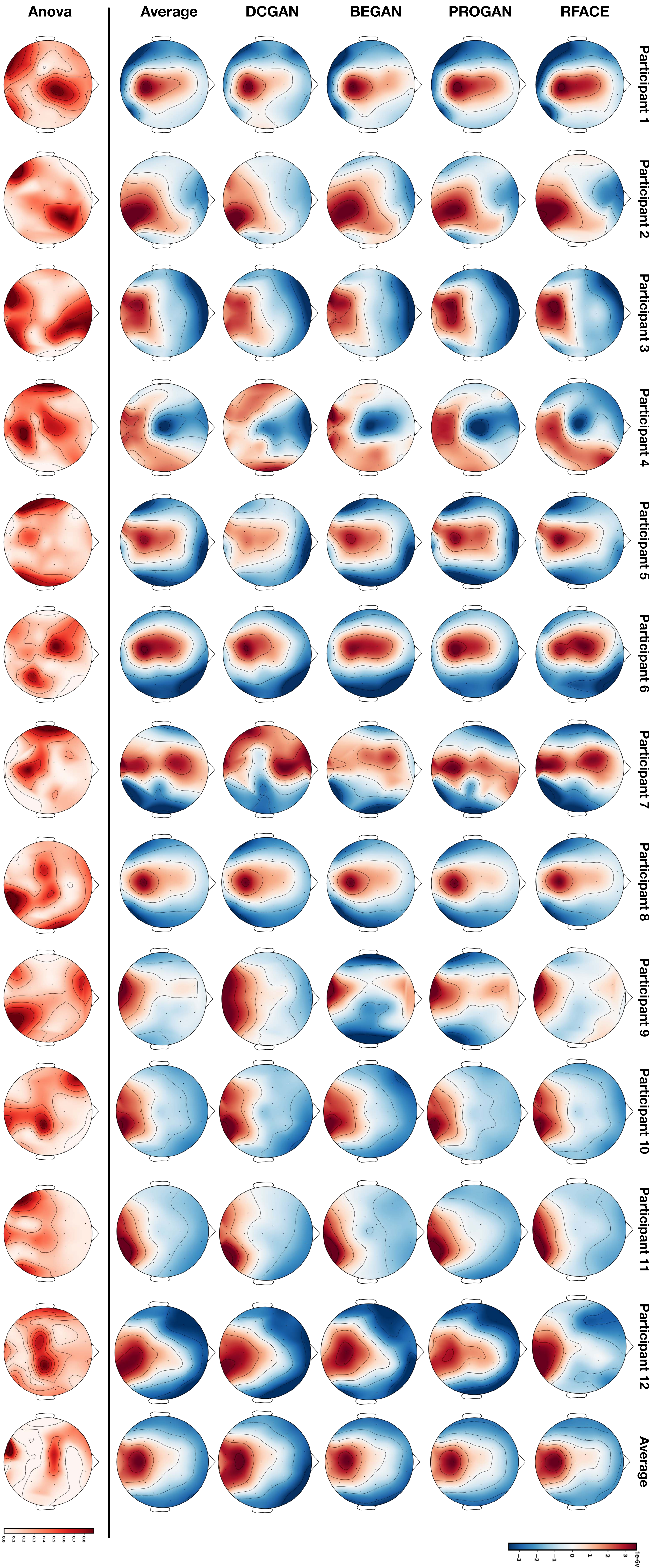}
	\caption{Averaged P300 topography of each participant for each category. F-values from an ANOVA test were computed for each channel across four categories. Topography is created at the optimal P300 time index for each participants which is demonstrated in  \citep{wang2018spatial}.} \label{fig-RSVP-P300-topo}
\end{center}
\end{figure}

Figure~\ref{fig-RSVP-P300-topo}
shows topographical plots (of averaged ERP activity) for the different image categories for each participant and for an average across participants. This demonstrates that the spatial topography of P300-related activity varies across participants. It is for this reason that we use the LDA beamformer approach to reconstruct the source P300 for each participant in this study (so as to eliminate erroneous measurement of the P300 by using a specific common channel). We also show a topographic representation of F-values from an ANOVA test that assesses statistical differences between the means of the four categories (one ANOVA for each channel). Larger F-values indicate a larger statistical effect when examining reconstructed P300 values across the four categories for a participant. It can be seen that spatial locations with high F-values are closely aligned to the P300's spatial topography.

We also show the Neuroscore for each participant in the study (for each GAN) in Table~\ref{neural-score}. A higher Neuroscore indicates better performance of a GAN. Ranking the performance of GANs by Neuroscore we  see: PROGAN $>$ BEGAN $>$ DCGAN, which is consistent with human judgment in the BE task. 
\begin{table}[!htbp]
	\centering
	\begin{tabular}{|c||c|c|c||c|}
		\hline
		{\textbf{\textit{ID}}}& {\textbf{\textit{DCGAN}}} & {\textbf{\textit{BEGAN}}} & {\textbf{\textit{PROGAN}}} & {\textbf{\textit{RFACE}}}  \\
		\hline \hline
		{1} &{0.577} & {0.668} & {0.685}  & {0.641}\\
		\hline
		{2} &{0.613} & {0.769}  & {0.939} & {0.820}\\
		\hline
		{3} &{0.446} & {0.630} & {0.689}  & {0.591}\\
		\hline
		{4} &{0.432} & {0.576}  & {0.974} & {0.930}\\
		\hline
		{5} &{0.658} & {0.907}  & {0.938} & {0.722}\\
		\hline
		{6} &{0.603} & {0.774}  & {0.964} & {0.811}\\
		\hline
		{7} &{0.462} & {0.584}  & {0.856} & {0.812}\\
		\hline
		{8} &{0.824} & {0.838}  & {0.882} & {0.789}\\
		\hline
		{9} &{0.683} & {0.722}  & {0.911} & {0.908}\\
		\hline
		{10} & {0.637} & {0.643} & {0.962} & {0.825} \\ 
		\hline
		{11} & {0.419} & {0.350} & {0.425} & {0.447} \\ 
		\hline
		{12} & {0.646} & {0.654} & {0.819} & {0.784} \\ 
		\hline \hline
		{Mean} &{\textbf{0.583}} & {0.676}  & {\textbf{0.837}} & {0.757}\\
		\hline
	\end{tabular}
	\caption{Computed Neuroscore for each participant for each category. Higher score indicates better performance of GAN.}
	\label{neural-score}
\end{table} 

\begin{figure}[!htbp] 
\begin{center}
	\includegraphics[scale=0.6]{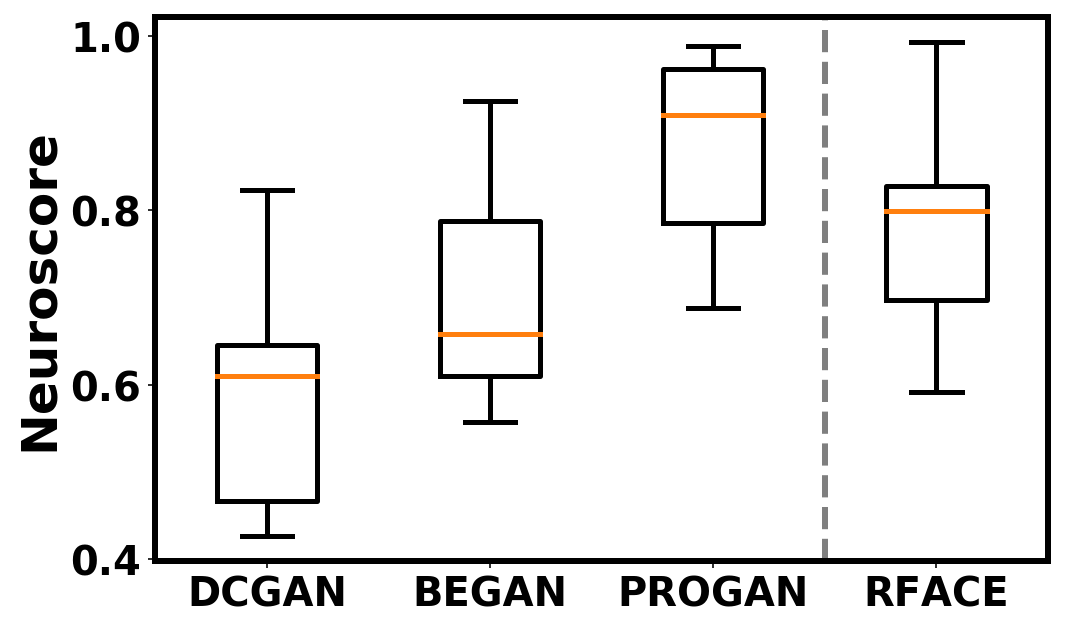}
	\caption{Box plot of Neuroscore for each image category across 12 participants.}
	\label{fig-ns-box}
\end{center}
\end{figure}

Figure~\ref{fig-ns-box} summarizes the details from Table~\ref{neural-score}. The median values of the Neuroscore for each category across participants give the same rank as the mean value in Table~\ref{neural-score}.

\begin{figure}[!htbp]
    \centering
    \includegraphics[width=\textwidth]{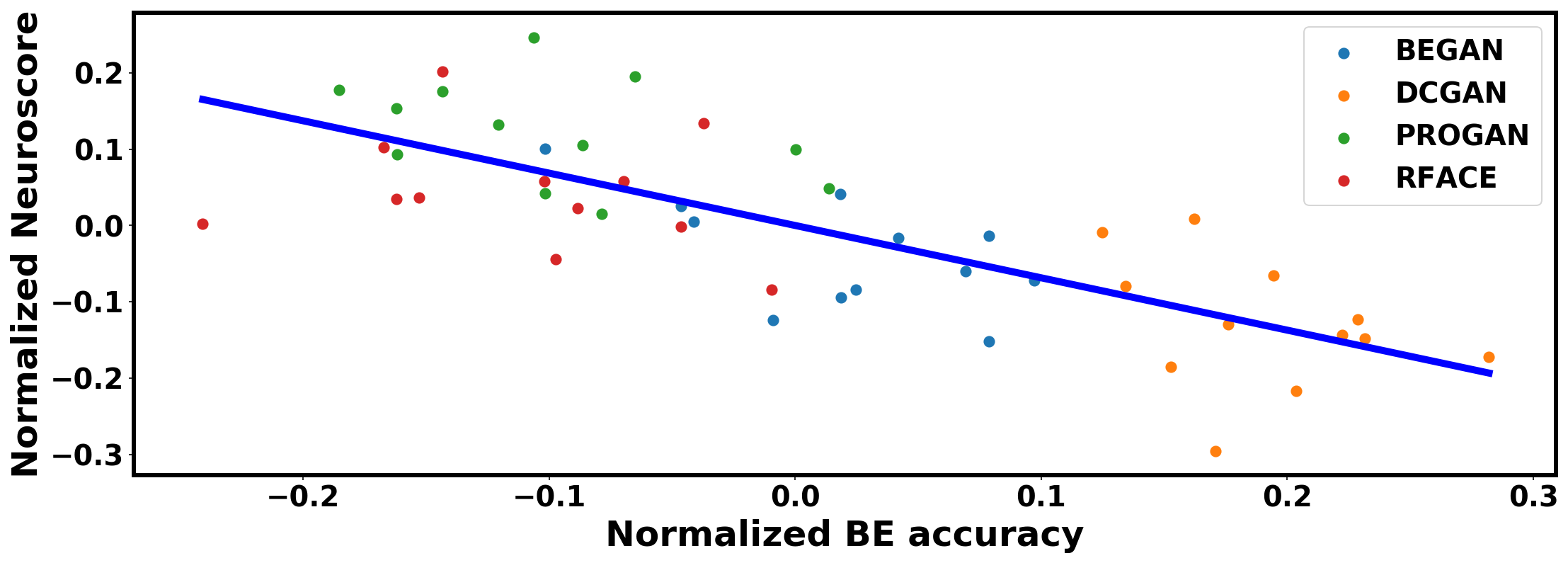}
    \caption{Correlation between Neuroscore and BE accuracy. Neuroscore and BE are both mean centered within each participant.}
    \label{fig-scatter-plot}
\end{figure}

From the averaged subtracted values (on a per-participant basis) of the Neuroscore and BE accuracies, it can be seen that the Neuroscore is correlated with the BE accuracy (human judgment) i.e., PROGAN $>$ BEGAN $>$ DCGAN (see Fig.~\ref{fig-scatter-plot}). 

In order to statistically measure this correlative relationship, we calculated the Pearson correlation coefficient and p-value (two-tailed) between Neuroscore and BE accuracy and found ($\mathbf{r(48)=-0.767, p=2.089e-10}$)\footnote{We also did the Pearson statistical test and bootstrap on the correlation between Neuroscore and BE accuracy only for GANs i.e., DCGAN, BEGAN and PROGAN. Pearson statistic is (\textbf{r(36)=-0.827, p=4.766e-10}) and the bootstrapped \textbf{p $\leq$ 0.0001}.}.

We used a bootstrap procedure \citep{efron1994introduction, 10.3389/fpsyg.2017.00456} to validate our Pearson correlation coefficient test since aggregating repeated measurements for participants (i.e., treating DCGAN, BEGAN, PROGAN and RFACE measurements as being independent) like this results in a violation of assumptions for our statistical test (violation of independence). Using a bootstrap procedure with our correlation measure allows us to sidestep this violation of assumptions and still obtain a reliable statistic. We do this by repeatedly randomly shuffling the BE accuracy values and Neuroscore (within each participant) and then applying a Pearson correlation coefficient test. After following this process 10,000 times, we count how many p-values calculated on randomly shuffled values (using within-participant shuffling) ($i$) are smaller than the original p-value (where within-participant shuffling is not applied). $\frac{i}{10000}$ now becomes the bootstrapped Pearson p-value i.e., it estimates the probability of getting the calculated p-value by chance. For the Pearson correlation coefficient test, this strongly supports the interpretation that our Neuroscore is predictive of human judgment. Due to time-based constraints in running the bootstrap procedure, we stopped at 10,000 iterations. This is consistent with our hypothesis that higher Neuroscore indicates better GAN models which is also indicated by lower BE accuracy. The bootstrapped p-value for the Pearson correlation coefficient test is significant ($\mathbf{p \leq 0.0001}$), which means that it is unlikely we have obtained these correlation results by chance\footnote{Without per-participant mean subtraction, the Pearson correlation statistic is ($\mathbf{r(48)=-0.556, p=4.038e-05}$) and the bootstrapped $\mathbf{ p \leq 0.0001}$.}.

It is notable that PROGAN achieved a higher Neuroscore than RFACE. There are differences between the RFACE and GAN generated images that are likely impacting the P300 amplitudes for the RFACE images. In the RFACE images, there are a wide range of background textures (e.g. sky, sea and indoor environments) that are not present in the GAN generated images. The GAN generated images tend to have homogeneous backgrounds, where in most cases they are almost monochromatic and/or out of focus. Furthermore, the RFACE images contain a greater variety of other artefacts (e.g., jewellery) that tend not to be discernibly reproduced by the GANs. The lower Neuroscore for RFACE (i.e., RFACE $<$ PROGAN) images is likely a result of these non-task related visual components in the RFACE images increasing the discrimination difficulty. It is known that increasing task difficulty results in a diminished P300 amplitude~\citep{kim2008influence}. For instance, increasing the amount of visual distractors in an image in a target detection task reduces the P300 amplitude~\citep{luck1990electrophysiological}. A further contributing factor may be the stereotyped visual structure of the GAN images (i.e., a face with a bland background), which facilitates the GAN images to be detected more easily in the fast RSVP paradigm used. From the human assessment results in the previous section, it can be seen that participants find the PROGAN output quite convincing, rating faces produced by the GAN similarly in accuracy as the RFACE images.

\subsection{Comparison to other evaluation metrics}
Three traditional methods are also employed to evaluate the GANs used in this study. Table~\ref{traditional-method-score}
shows the scores from the three traditional metrics, Neuroscore and human judgment for three GANs.  
\begin{table}[!htbp] 
	\centering
	\begin{tabular}{|c|c|c|c|}
		\hline
		{\textbf{Methods}}& {\textbf{DCGAN}} & {\textbf{BEGAN}} & {\textbf{PROGAN}}\\
		\hline \hline
		{1/IS} &{\textcolor{red}{0.44}} & {\textcolor{red}{0.57}} & {0.42} \\
		\hline
		{MMD} &{\textcolor{red}{0.22}} & {\textcolor{red}{0.29}}  & {0.12}\\
		\hline
		{FID} &{\textcolor{red}{63.29}} & {\textcolor{red}{83.38}} & {34.10}\\
		\hline
		{1/Neuroscore} &{1.715} & {1.479} & {1.195}\\
		\hline
		{Human} &{\textbf{0.995}} & {\textbf{0.824}} & {\textbf{0.705}} \\
		\hline
	\end{tabular}
	\caption{ Score comparison for each GAN category. Lower score indicates better performance of GAN.}
	\label{traditional-method-score}
\end{table}
To be consistent with other metrics (smaller score indicates better GAN performance), we use 1/Neuroscore for comparison. It can be seen that all three methods are consistent with each other and they rank the GANs in the same order of PROGAN, DCGAN and BEGAN from high  to low performance. By comparing the three traditional evaluation metrics to the human, it can be seen that they are not consistent with human judgment of GAN performance. It should be remembered that Inception Score is able to measure the quality of the generated images \citep{salimans2016improved} while the other two methods cannot do so. However, Inception Score still rates DCGAN as outperforming BEGAN. Our proposed Neuroscore is consistent with human judgment.

\section{Discussion}
We have compared human assessment with three representative quantitative metrics and used these for comparison with our proposed neural scoring approach. In short, our Neuroscore conveys a measure of the visual quality of facial images generated from GANs. This is based on our hypothesis that a generated image which looks more like a real face image will elicit a larger \textit{reconstructed averaged P300 amplitude} in an RSVP task. Although the other three traditional evaluation methods do provide insight into several aspects of GAN performance, we study their effectiveness from a visual image quality perspective only as this is the focus of our work. The results are compelling in their demonstration that the proposed Neuroscore is better correlated with human judgment than any of the three quantitative metrics. This is important as an evaluation of the visual quality of a generated image is useful in understanding performance characteristics of specific GAN designs and training data sets. The method proposed can meet this need and is independent of any data modelling assumptions. In contrast, conventional quantitative metrics may fail in this regard. 

For example, Inception Score is a model-based evaluation method and the model is very sensitive to adversarial samples as shown in \citep{kurakin2016adversarial}. Inception Score will also produce a very high score if the generated images are produced using adversarial training \citep{barratt2018note}. Our Neuroscore approach would not be compromised with such images in comparison. It is worth noting that compared with MMD and FID, both Inception Score and our Neuroscore provide a potentially good way of comparing the visual quality between generated images and real images i.e., Inception Score and Neuroscore may give higher score for the generated image that has better visual quality than the real image. Inception Score, however, unlike the neural scoring approach is not able to improve on the ranking of the three GANs compared to MMD or FID. 

As mentioned earlier, more realistic GANs will produce a higher Neuroscore. This is because Neuroscore is sensitive to different stimulus processing requirements for different types of GANs i.e., the larger averaged single trial P300 amplitudes for GANs  reflect properties related to different stimulus information processing requirements \citep{sur2009event}. It is also worth commenting that while GANs for generating facial images are explored in this study, our approach could be used for other types of generated images because the P300 ERP can be elicited using a wide variety of significantly different visual stimuli e.g., Neuroscore may be applicable in the evaluation of GANs in bedroom image generation \citep{ karras2017progressive, mao2017least, radford2015unsupervised, yu2015lsun}.

The work presented here focuses on evaluating image visual quality only. Consequently there are some limitations when using the Neuroscore to evaluate GANs in this way. Overfitting, mode dropping and mode collapsing are very important aspects of GAN performance and most quantitative methods are able to assess these in some way. However for these broader assessments, we can augment quantitative methods with our Neuroscore to gain a better assessment of overall GAN performance. In reality, choosing the appropriate evaluation metric for GANs depends on the application and which type of problem is being addressed by the GAN. If the goal of the GAN application is the generation of high visual quality images, e.g. super resolution image reconstruction, a qualitative metric is preferred in that case. If the GAN is to be trained to capture the categories of large image datasets, a quantitative metric would be a better choice. Therefore the inclusion of a neural scoring approach as we have demonstrated should be considered in the context of the application's requirements.  

Neuroscore is produced from human EEG signals and directly reflects  human perception and neural processes. Compared to human judgment on images generated from GANs, our paradigm has several advantages as follows. Firstly, it is much faster than human judgment as a rapid image stream is presented to participants as part of the RSVP protocol. Traditional human judgment approaches entails the evaluation of images one-by-one whereas our paradigm supports batch evaluation of images. Secondly, as the EEG recorded corresponds to individual images, the method allows the tracking of image quality at the level of the individual image rather than the aggregated quality of a group of images. Thirdly, Neuroscore produces a continuous value while human judgment is binary (``real'' or ``fake'').  Finally, it is possible to use EEG signals such as P300 as supervised information for improving training of GANs in the future.

In this work, we focus on the  evaluation of images generated from GANs. However, time series evaluation of GANs is even more challenging and even less discussed in the literature. We believe that our paradigm may extend to use the auditory BCI~\citep{cai2015brain} for auditory evaluation for GANs in the future.

\section{Conclusion}
We have conducted a comprehensive comparison between human assessments and three quantitative metrics for the comparison of image quality in the specific GAN application of facial image synthesis. We proposed and assessed a neural interfacing approach in which a Neuroscore is introduced as an alternative evaluation of GANs in terms of image visual quality. We interpret our results to conclude that Neuroscore is more consistent with assessments made by humans when compared to the three established quantitative metrics and we show that the correlation between our Neuroscore and human judgment is not produced by chance i.e., p $\leq$ 0.0001. We believe that our proposed neuro-AI interface based on a rapid serial visual presentation approach is more efficient and less prone to error compared to conventional human annotation. Consequently we suggest that approaches using such neural signals may complement or for some specific applications, replace, conventional metrics for evaluation of GAN performance.

\section*{Conflict of Interest}
Zhengwei Wang declares that he has no conflict of interest. Graham Healy declares that he has no conflict of interest. Alan F. Smeaton declares that he has no conflict of interest. Tom\'as E. Ward declares that he has no conflict of interest.

\section*{Ethical Approval}
All procedures performed in studies involving human participants were in accordance with the ethical standards of the institutional and/or national research committee and with the 1964 Helsinki declaration and its later amendments or comparable ethical standards.  Formal approval for this work was given from Dublin City University Research Ethics Committee (REC/2018/115).

\begin{acknowledgements}
This work is funded as part of the Insight Centre for Data Analytics which is supported by Science Foundation Ireland under Grant Number SFI/12/RC/2289.
\end{acknowledgements}


%
%

\bibliographystyle{spbasic}
\bibliography{ref.bib}

\end{document}